\documentclass[a4paper, 10pt, conference]{ieeeconf}     

\IEEEoverridecommandlockouts                              

\usepackage{times}
\usepackage{epsfig}
\usepackage{amsmath}
\usepackage{gensymb}
\usepackage{graphicx}
\usepackage{float}
\usepackage{mathtools,amssymb}
\usepackage{algorithm, algpseudocode, float}
\usepackage{amsfonts}
\usepackage[hyphens]{url}
\usepackage{textcomp}
\usepackage{diagbox}
\usepackage{array}
\usepackage[noadjust]{cite}
\usepackage[ruled,vlined, algo2e]{algorithm2e}

\newcolumntype{P}[1]{>{\centering\arraybackslash}p{#1}}

\algnewcommand\algorithmicforeach{\textbf{for each}}
\algdef{S}[FOR]{ForEach}[1]{\algorithmicforeach\ #1\ \algorithmicdo}

\title{\LARGE \bf
	CurbScan: Curb Detection and Tracking Using Multi-Sensor Fusion
}

\author{Iljoo Baek$^1$, Tzu-Chieh Tai$^1$, Manoj Mohan Bhat$^1$, Karun Ellango$^2$ \\
Tarang Shah$^1$, Kamal Fuseini$^1$, Ragunathan (Raj) Rajkumar$^1$ \\
\emph{$^1$Carnegie Mellon University, $^2$The Bronx High School of Science}
}

\begin{document}
	
\maketitle
\thispagestyle{empty}
\pagestyle{empty}

\begin{abstract}
Reliable curb detection is critical for safe autonomous driving in urban contexts. Curb detection and tracking are also useful in vehicle localization and path planning. Past work utilized a 3D LiDAR sensor to determine accurate distance information and the geometric attributes of curbs. However, such an approach requires dense point cloud data and is also vulnerable to false positives from obstacles present on both road and off-road areas. In this paper, we propose an approach to detect and track curbs by fusing together data from multiple sensors: sparse LiDAR data, a mono camera and low-cost ultrasonic sensors. The detection algorithm is based on a single 3D LiDAR and a mono camera sensor used to detect candidate curb features and it effectively removes false positives arising from surrounding static and moving obstacles. The detection accuracy of the tracking algorithm is boosted by using Kalman filter-based prediction and fusion with lateral distance information from low-cost ultrasonic sensors. We next propose a line-fitting algorithm that yields robust results for curb locations. Finally, we demonstrate the practical feasibility of our solution by testing in different road environments and evaluating our implementation in a real vehicle\footnote{Demo video clips demonstrating our algorithm have been uploaded to

Youtube: https://www.youtube.com/watch?v=w5MwsdWhcy4, \\ https://www.youtube.com/watch?v=Gd506RklfG8.}. Our algorithm maintains over 90\% accuracy within 4.5-22 meters and 0-14 meters for the KITTI dataset and our dataset respectively, and its average processing time per frame is approximately 10 ms on Intel i7 x86 and 100ms on NVIDIA Xavier board.

\end{abstract}

\section{INTRODUCTION}
Complex and dynamic road environments pose substantial challenges in guaranteeing safety for vehicle localization and path planning. 
For example, moving or parked vehicles, construction zones and urban roads change the geometric characteristics of the environment. This in turn can lead to large and unsafe localization and path planning errors.
A curb can be deemed as a strong road feature that delimits the road boundaries, and can provide rich information for use in vehicle positioning~\cite{suhr2016sensor, hata2017monte}.
Therefore, the ability to detect road curbs is paramount to ensure the safe navigation of autonomous vehicles. A 3D LiDAR sensor has been used in recent work to determine accurate distance information and the geometric characteristics of curbs.
Small-form factor LiDAR sensors such as the Velodyne VLP-16 have been used to reduce sensor costs and to yield a better appearance. However, the lower resolution of the LiDAR creates the need to detect curbs using only sparse points on a curb.
In this paper, we present a solution based on sensor fusion to detect and track road curbs using sparse 3D LiDAR, a mono camera, and ultrasonic sensors in real time. This solution was developed for use in Carnegie Mellon University's autonomous driving research vehicle.

\subsection{Related work}

There have been numerous studies in the past aimed at detecting curbs and road boundaries. These include single 2D-LiDAR sensor-based approaches used by~\cite{demir2017adaptive, han2014road, liu2013new, kang2012LiDAR, wijesoma2004road}. In this approach, curb features are extracted and road boundaries use a fixed pitch angle, a series of depth and angle measurements for each LiDAR, and ego-vehicle movement information. Since a 2D-LiDAR scanner provides sparse points per frame, these solutions work in real-time without hardware accelerators. However, the sparse 2D scan data can be easily affected by noise in a complex urban road environment.
Another approach utilizes a single 3D LiDAR sensor to extract geometric characteristics from a very dense and high-resolution point cloud with a larger coverage area~\cite{wang2019point, sun20193d, zhang2018road, zai20173, wang2015road, chen2015velodyne, zhang2015real, guan2014automated}. This approach first extracts curb or road boundary features using the geometric characteristics of the LiDAR point distributions such as the height change or continuity. Then, the extracted points are classified as candidate curb points. After filtering out noise points, regression algorithms such as RANSAC are applied to fit the points to curve lines. The curve points can be predicted with a Kalman filter. The disadvantage of this approach is that it does not consider the presence of dynamic obstacles or occlusions in the environment, which can adversely affect curb detection performance. 
In order to overcome this problem, Hata et al.~\cite{hata2015feature, hata2014robust} applied least trimmed squares (LTS) regression~\cite{rousseeuw2006computing} as the fitting method for the regression filter. Xu et al.~\cite{xu2016road} connect missing curb lines due to small occlusions by obstacles using the optimal path produced by a least-cost path model (LCPM). However, these methods do not work well on false positive curb points on large obstacles such as a wall or a fence. A new approach~\cite{xu2019real} utilizes road consistency and obstacle maps by rearranging dense point cloud data. These maps provide the spatial relationship and height difference between the road and off-road, regions, which help to segment road region from unstructured rural scenes.
There have been other techniques to detect road boundaries using LiDAR and other sensors~\cite{gezero2019automated, huang2017practical, tan2014robust, li2013sensor}.
Global Navigation Satellite Systems (GNSS) have been used to remove the presence of obstacles in the road area or non-ground points in predicting curb trajectories~\cite{huang2017practical, gezero2019automated}.
Li et al.~\cite{li2013sensor} propose a feature-level fusion method to select the best road boundary features from multiple 2D LiDAR and camera sensors using prior knowledge. 
Another approach~\cite{tan2014robust} is to recover a dense depth image of the captured scene using 3D LiDAR data and a high-resolution image. Then, they calculate a normal image from the depth image and detect curb point features by using normal patterns based on the curb’s geometric property. Dense LiDAR data and accuate calibration between sensors are core to the algorithm.

In summary, the challenges in prior curb detection studies using 3D
LiDAR point clouds are as follows.
First, they require dense LiDAR points to detect the geometric characteristics of curbs. Sparse LiDAR data can degrade their detection performance significantly. 
Second, false-positive points on large obstacles such as a wall or fence can be misclassified as curbs. 
Third, in the evaluation of curb detection, they do not consider longitudinal accuracy, which can provide confidence in localization for autonomous vehicles.

\subsection{Our Contributions}

Previous contributions have helped us significantly in shaping our research. We have adopted the filter-based algorithms~\cite{gezero2019automated, hata2014robust, xu2016road} to extract candidate curb features from LiDAR point data. Then, the candidate points are filtered using a virtual scan algorithm~\cite{mengwen2016robust} and an image-based object detector~\cite{ssd-mobileNet-tensorflow} to remove the points belonging to general obstacles for removing false positives. 
We have also extended the Kalman filter-based tracking approach~\cite{wang2019point, wang2015road} to overcome missing detections due to temporary occlusions.

The main contributions of our paper are as follows:

\begin{enumerate}

\item We present a real-time solution to effectively detect road curbs in an urban environment using sparse 3D LiDAR data such as Velodyne VLP-16.

\item We provide an efficient masking method to remove false positives due to dynamic or static obstacles in on-road and off-road areas.

\item We propose a practical and deterministic curve-fitting method that outperforms traditional non-deterministic algorithms such as RANSAC.

\item We take into account the longitudinal accuracy of curb detection, which provides confidence in localization estimation.

\item We demonstrate the usefulness of integrating low-cost distance sensors such as ultrasonic sensors in road curb and boundary detection.

\end{enumerate}

\section{BACKGROUND}

\subsection{LiDAR}
Light Detection and Ranging (LiDAR) technology has been widely used as a reliable and informative sensor on autonomous vehicles in recent years. Compared to other sensors like cameras, 3D LiDARs provide accurate depth and position information of the objects in the environment. Much research has been focusing on LiDARs with 64 or 32 scan lines, since they provide point cloud data with higher density. However, the cost of these hi-res LiDARs is a big hurdle for commercialization and aesthetic appearance. In our study, we choose the compact Velodyne VLP-16, which provides only 16 scan lines and around 30000 maximum points per frame when running at 10Hz. 

\subsection{Ultrasonic Sensor}

Ultrasonic sensors are time-of-flight sensors that rely on sound waves for distance estimation. These sensors include a transmitter and receiver which are used to estimate the time-of-flight of an emitted sound wave. This time of flight helps determine the distance of the object from the sensor. 
The major drawbacks of this sensor include that it only provides a one-dimensional(1-D) output of the distance. Also, changes in the distance cause noisy sensor outputs. However, ultrasonic sensors are much faster than other sensors. They also have the benefit of being inexpensive and are widely available.




\begin{figure}[]
\centerline{\includegraphics[width=0.8\columnwidth]{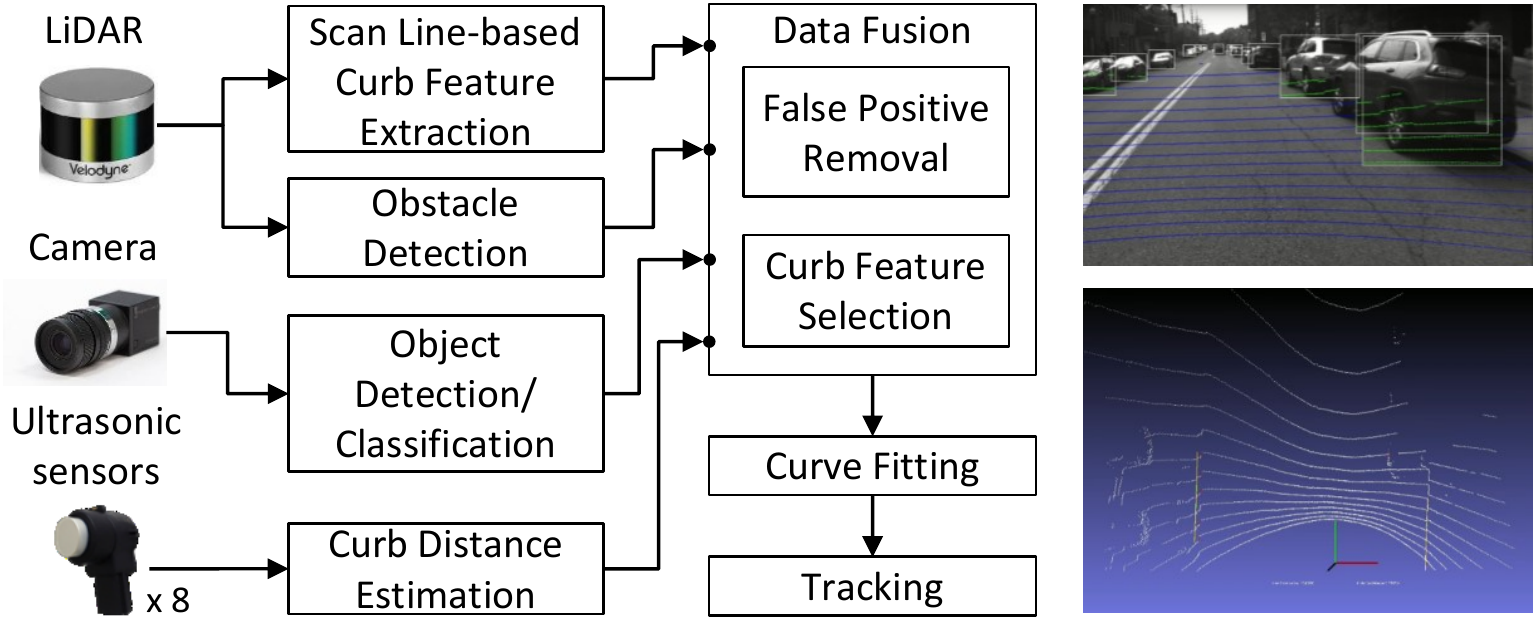}}
\caption{Left: Proposed curb detection architecture- LiDAR point cloud is the input for curb feature extraction. LiDAR-based obstacle detection(VScan) and Camera-based object detector remove false-positive curb points from obstacles. Ultrasonic sensor-based curb detection is fused to compensate sparse curb points from LiDAR and increase the accuracy of the curve fitting. Finally, the tracking module tracks the curb lines occluded by nearby objects. Right: Example visualization of raw LiDAR points and detected curb lines in the LiDAR and image planes with the CurbScan system.}
\label{fig_overall_architecture}
\end{figure}

\begin{figure}[b]
\centerline{\includegraphics[width=0.5\columnwidth]{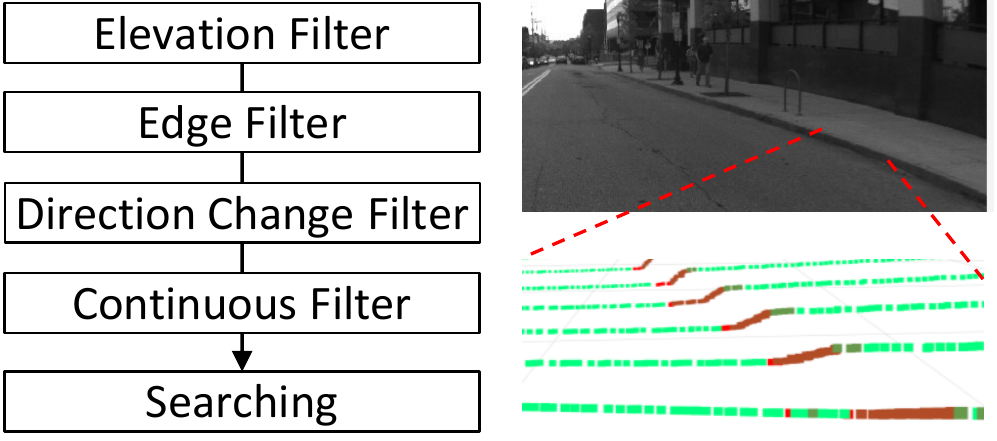}}
\caption{Left: Multiple filters are designed to extract curb features from sparse LiDAR data effectively. The Searching algorithm determines candidate curb points from the extracted curb features. Right: The different colors of the LiDAR points represent the results from various filters.}
\label{fig_LiDAR_curb_feature}
\end{figure}

\section{CurbScan Architecture and Algorithms}
\label{architecture_algorithm}

Figure~\ref{fig_overall_architecture} shows the pipeline for our curb detection and tracking modules. The function and design of these modules are explained in the following sections.

\subsection{LiDAR-based detection}


\smallskip
\textbf{Filter-based curb detection}

To utilize point cloud data returned from the Velodyne VLP-16 LiDAR, data preprocessing is necessary. The VLP-16 LiDAR emits 16 separate lasers and performs a $360^{\circ}$ scan to return angular and distance information about objects in the surroundings. The raw data is represented in spherical coordinates with $\{$radial distance, azimuth, vertical angle$\}$ for each point. We convert the data to Cartesian coordinates.

Figure \ref{fig_LiDAR_curb_feature} shows our processing pipeline for LiDAR point cloud. Several filters are applied to get the defined features such as \emph{elevation filter}, \emph{edge filter}, \emph{direction change filter}, and \emph{continuous filter}. Possible curb points are determined by a search algorithm based on these features.

\smallskip
\textbf{Filter-based feature extraction}

Several filters are applied to the point cloud data to help determine the possible curb points. To find the road boundary points like curbs, several geometric features are defined to find the candidate feature points. These feature points are followed by scan-based searching to determine the curb points.

Since each laser beam from VLP-16 continuously fires at a fixed vertical angle, all returned ground points form a circle on a flat surface assuming the LiDAR is parallel to the surface. For non-level orientation, the scan lines would be ellipses or parabolas depending on the LiDAR orientation. In any case, the curve is smooth when the laser does not hit any obstacle with an elevation change. We utilize this property to detect the elevation deviations where curbs are present. Algorithm 1 presents our direction change filter. If focusing on a small number of consecutive laser points from $P_{i-k}$ to $P_{i+k}$, we define the angle $\theta_{i}$ at center point $P_i$ as 

\smallskip
\smallskip
\scalebox{0.8}{
\begin{minipage}{0.97\linewidth}

\begin{equation}
\theta_{i} = \arccos{(\frac{Avg(left)\cdot Avg(right)}{||Avg(left)||Avg(right)|} )}
\end{equation}

\end{minipage}%
}

\smallskip
\smallskip
$Avg(left)$ and $Avg(right)$ are defined as

\smallskip
\scalebox{0.8}{
\begin{minipage}{0.97\linewidth}

\begin{equation}
Avg(left) = \sum_{j=i-1}^{i-k}\overrightarrow{{P_{j}P_{i}}}
\end{equation} 
\begin{equation}
Avg(right) = \sum_{j=i+1}^{i+k}\overrightarrow{{P_{j}P_{i}}}
\end{equation} 

\end{minipage}%
}

$\overrightarrow{{P_{j}P_{i}}}$ represents the vector from $P_{i}$ to $P_{j}$ and $Avg(left)$ is the average direction from $P$. By taking the average of k points, noises is reduced. $\theta_{i}$ can be calculated after having two direction vectors $Avg(left)$ and $Avg(right)$. If $\theta_{i}$ is lower than a threshold and also a local minimum, the point is marked as a point where there is a change in elevation.

\begin{algorithm}[H]
\scalebox{0.75}{
\begin{minipage}{1.3\linewidth}
\SetAlgoLined
    \SetKwInOut{Input}{input}
    \SetKwInOut{Output}{output}
    \Input{$P$: A vector of point cloud with size $N$ ordered by scan number and azimuth}
    \Output{$res$: A vector of Boolean for each point}
    \smallskip
    \For{$i = 1, 2, ..., N$}{
    $\theta_{i} = \arccos{(\frac{Avg(left)\cdot         Avg(right)}{||Avg(left)||Avg(right)|} )}$\;
    where $Avg(left) = \sum_{j=i-1}^{i-k}\overrightarrow{{P_{j}P_{i}}}$ and $Avg(right) = \sum_{j=i+1}^{i+k}\overrightarrow{{P_{j}P_{i}}}$
    }
    \smallskip
    \For{$i = 1, 2, ..., N$}{
    \eIf{$\theta_{i} < thres$ and $\theta_i == localMin$}{
        $res[i] = true$
    }{
        $res[i] = false$
    }
    }
\end{minipage}%
}
    \caption{Direction Change Filter}
\end{algorithm}

\begin{algorithm}[H]
\scalebox{0.75}{
\begin{minipage}{0.95\linewidth}
\SetAlgoLined
    \SetKwInOut{Input}{input}
    \SetKwInOut{Output}{output}
    \Input{$P$: A vector of point cloud with size $N$ ordered by scan number and azimuth}
    \Output{$res$: A vector of Boolean for each point} \smallskip
    \For{$i = 1, 2, ..., N$}{
    \eIf{$P[i].z-P[i-1].z > thres$}{
        $res[i] = true$
    }{
        $res[i] = false$
    }
    }
\end{minipage}%
}
\caption{Elevation Filter}
\end{algorithm}

The elevation change of consecutive points in each scan line is another feature that gives some hints on finding potential curb points. Algorithm 2 lists the steps in our elevation filter. The filter calculates the height (z-axis) difference between the current and previous points in scan order. If the height difference is higher than a threshold, the point is classified as bieng elevated.

An edge filter, presented in Algorithm 3, next takes the result from the elevation filter as its input, and produces road edge information as its output, which are discussed in Algorithm 3. It takes into account \textit{n} consecutive points on the left and \textit{n} consecutive points on the right from the current point $i$ to determine if the current point is between a smooth surface and a rising edge.

A continuous filter, shown in Algorithm 4, calculates the distance between two consecutive points along each scan line. If the distance is higher than a threshold, both points are marked as discontinuous points.

\begin{algorithm}[H]
\scalebox{0.78}{
\begin{minipage}{1.25\linewidth}
\SetAlgoLined
    \SetKwInOut{Input}{input}
    \SetKwInOut{Output}{output}
    \Input{$isElevated$: A Boolean vector of whether the point is elevated along the scan direction}
    \Output{$isEdgeStart$: A Boolean vector of whether the point is between a smooth plane and a rising edge along scan direction
        \newline
        $isEdgeEnd$: A Boolean vector of whether the point is between rising edge and smooth plane along the scan direction
    }
    \smallskip
    \For{$i = 1, 2, ..., N$}{
    \If{$cntLeftElevated(i) < thres$ and $cntRightElevated(i) > thres$}{
        $isEdgeStart[i] = true$
    }
    \If{$cntLeftElevated(i) > thres$ and $cntRightElevated(i) < thres$}{
        $isEdgeEnd[i] = true$
    }
    }
\end{minipage}%
}
\caption{Edge Filter}
\end{algorithm}

\vspace{0.00mm}

\begin{algorithm}[H]
\scalebox{0.78}{
\begin{minipage}{0.97\linewidth}
\SetAlgoLined
    \SetKwInOut{Input}{input}
    \SetKwInOut{Output}{output}
    \Input{$P$: A vector of point cloud with size $N$ ordered by scan number and azimuth}
    \Output{$res$: A vector of Boolean for each point}
    \smallskip
    \For{$i = 1, 2, ..., {N-1}$}{
        \eIf{$sqrt(|P[i+1]-P[i]|^2) < thres$}{
            $res[i] = true$
        }{
            $res[i] = false$
        }
    }
\end{minipage}%
}
\caption{Continuous Filter}
\end{algorithm}

\smallskip
\subsection{Sensor Calibration}
\begin{figure}[]
\centerline{\includegraphics[width=1.0\columnwidth]{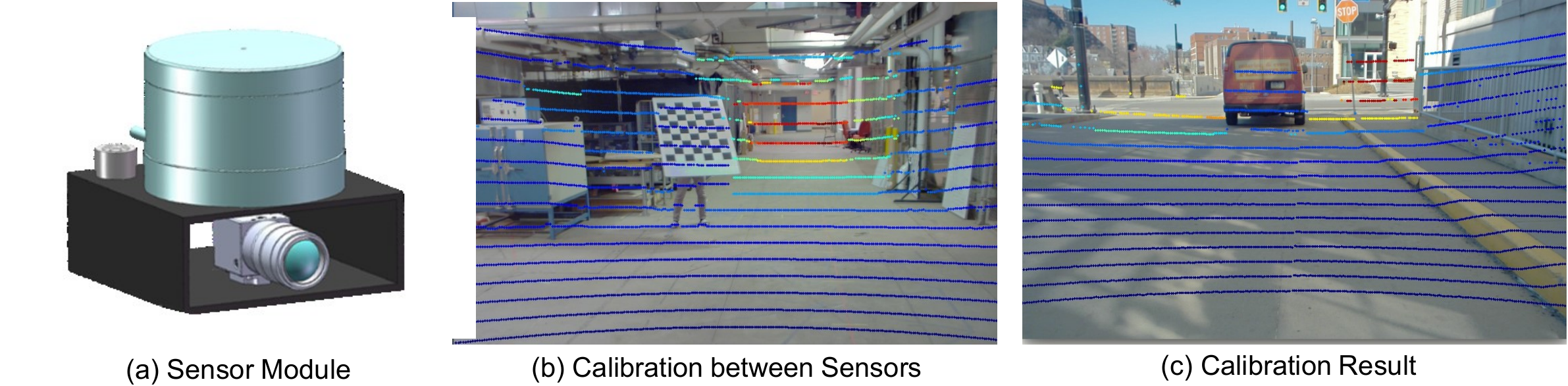}}
\caption{(a) LiDAR and Camera module mounted on our test vehicle. (b) Example scene with the calibration board (c) Projected point cloud from 3D LiDAR coordinate to 2D image plane after calibration.}
\label{fig_LiDAR_cam_calibration}
\end{figure}

Our LiDAR and monocular camera sensor pair need to be calibrated to correlate the information produced by each.
We use the Autoware software suite \cite{autoware} to generate and find these calibration parameters. The LiDAR and camera are mounted together as seen in Figure \ref{fig_LiDAR_cam_calibration}(a). Their relative positions and distance to each other are measured and input to Autoware to perform intrisic and extrinsic calibration. 

Intrinsic calibration is done by using the checkerboard in Figure \ref{fig_LiDAR_cam_calibration}(b) and defining the parameters of the checkerboard in Autoware. Then, the intrinsic parameters of the camera will be automatically generated by running the autoware\_camera\_calibration script based on the official ROS calibration tool.

Extrinsic calibration is done by clicking on corresponding points in the image and the point cloud. After correlating a few of the points manually by clicking on a pixel and the related LiDAR point, the transformations between the LiDAR and camera will be automatically generated and the camera-LiDAR calibration will be completed. Figure\ref{fig_LiDAR_cam_calibration}(c) shows re-projected LiDAR points on the corresponding image plane using this calibration.

\begin{figure}[]
\centerline{\includegraphics[width=0.6\columnwidth]{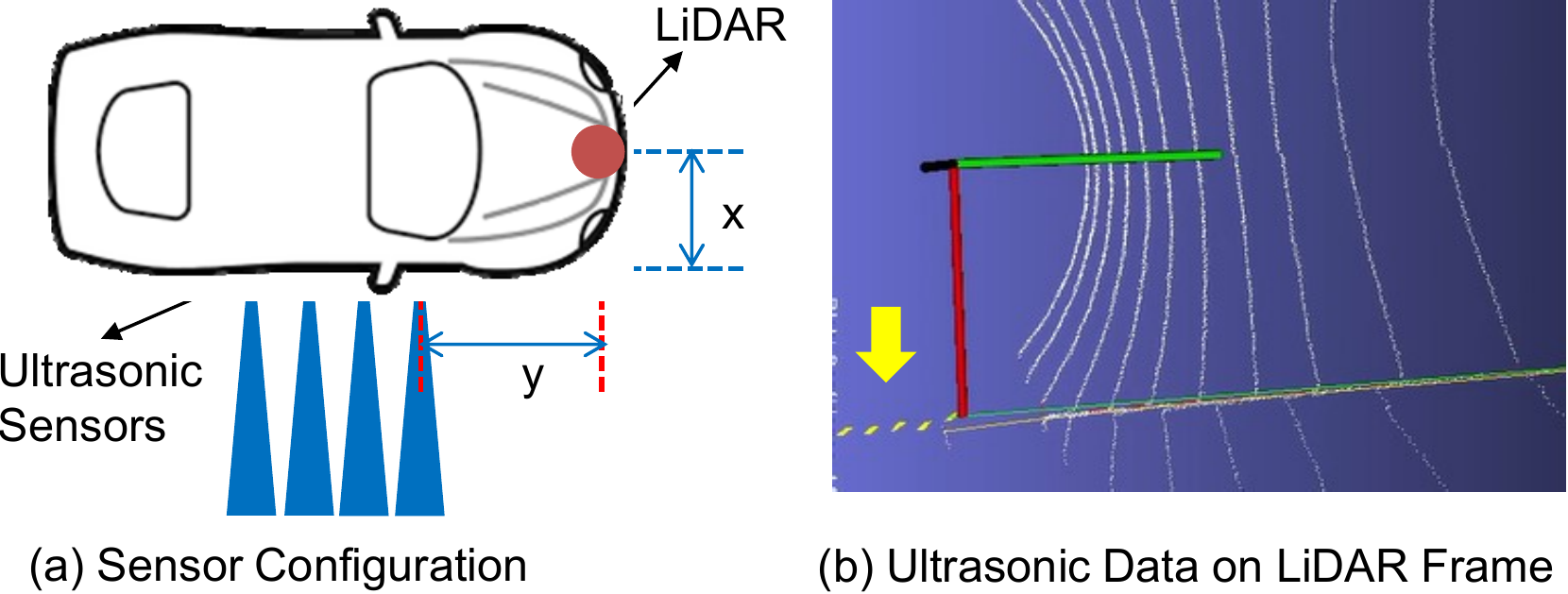}}
\caption{(a) Sensor configuration between ultrasonic sensors and LiDAR sensor. (b) Reprojected ultrasonic data to LiDAR Coordinate. The yellow arrow indicates the ultrasonic data.}
\label{fig_LiDAR_Ultrasonic_calibration}
\end{figure}

To transform the detected distances from our ultrasonic sensors to the LiDAR frame, we use a translational transform. We measure the offsets between the center of the LiDAR and each of the ultrasonic sensors as shown in Figure~\ref{fig_LiDAR_Ultrasonic_calibration}(a). For simplicity, we assume the distance predicted is along the axis of the ultrasonic sensor. Figure~\ref{fig_LiDAR_Ultrasonic_calibration}(b) shows re-projected ultrasonic points on the corresponding LiDAR plane. The z-axis values of the ultrasonic data correspond to the height of the sensor from the ground.

\subsection{False Postive Removal}

\begin{figure}[]
\centerline{\includegraphics[width=0.93\columnwidth]{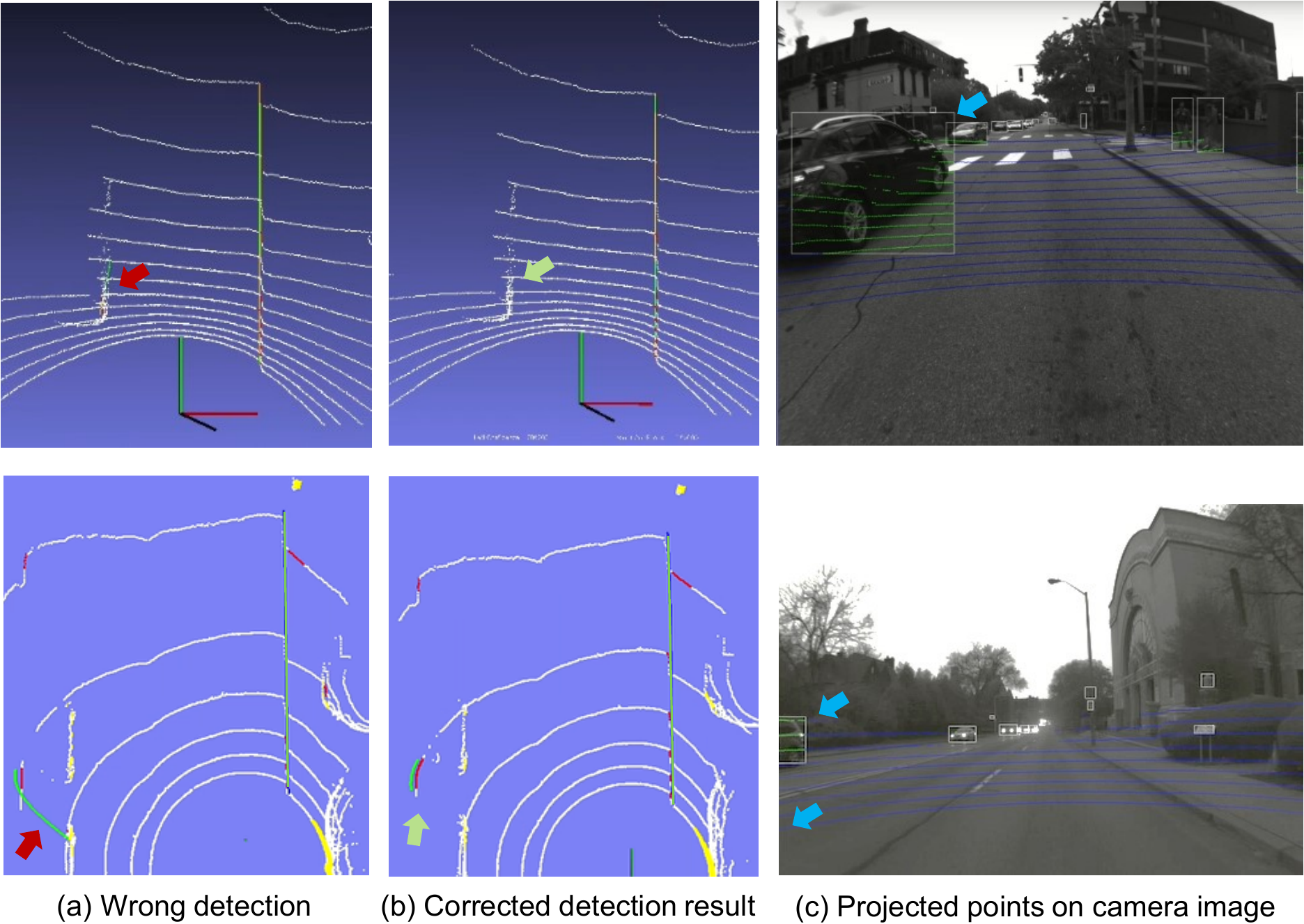}}
\caption{The top and bottom row pictures show an example case using Vision object detector-based and VScan-based masking to remove false-positive curb points on nearby cars, respectively.  (a) The red arrow indicates the false positive detection of the vehicle. (b) The green arrow shows that the false-positive curb line is removed and the correct curb is detected after masking. (c) Bounding boxes of detected objects from the object detector. The blue arrow indicates an object where the false-positive points belong to. Note that the 2nd blue arrow at the bottom picture indicates a car obstacle that is not in the field-of-view of the camera. VScan is helpful in this case.}
\label{fig_vision_masking}
\end{figure}

False positive curb features may be generated by our filter due to the presence of moving and stationary obstacles. We filter these false positives using an object detector (based on SSD MobileNetv1) \cite{ssd-mobileNet-tensorflow} and virtual scan \cite{mengwen2016robust} as shown in Figure \ref{fig_vision_masking}.

\smallskip
\textbf{Vision-based object detection}

SSD MobileNetv1 \cite{ssd-mobileNet-tensorflow} is a very effective neural network for recognizing objects with a camera. After the calibration step, where the LiDAR and camera, as seen in Figure \ref{fig_LiDAR_cam_calibration}(a), are cross-correlated, objects detected by SSD MobileNetv1 can be excluded from the generated boundary lines. The SSD MobileNetv1 first detects any potential objects and calculates the confidence interval of the object. Then, the camera image is projected to the LiDAR point cloud. If any boundaries generated by SSD MobileNetv1 intersect with the curb points returned after LiDAR processing, they are removed. This can be seen in Figure~\ref{fig_vision_masking}(a), where the points that represent the cars in Figure~\ref{fig_vision_masking}(c) that are also in the boundary points generated by LiDAR processing are included even though they should not be. However, after cross-correlating the data, the false boundary points generated by the cars are removed, yielding the proper curb boundaries, as seen in Figure~\ref{fig_vision_masking}(b).

\smallskip
\textbf{Fast Virtual Scan-based boundary detection}
We also use VScan (virtual scan)~\cite{mengwen2016robust} to remove false positives from the LiDAR point cloud that are not in the camera field-of-view and when the camera cannot perform well at night or rainy weather.
First, the 3D point cloud is compressed into a 2D plane to allow for faster operations with less computational cost. A basic VScan matrix is used to represent the point cloud around the vehicle. Then, Simultaneous Road Filtering and Obstacle Detection (SRFOD) is run on the matrix which when optimized with a Sorted Array-based Acceleration Method allows for robust, real time VScan generation. The SRFOD allows for the generation of "stixels" which are used to represent the height of an object. These stixels are then used for filtering out false positives by cross-correlating them with the set of boundary points generated and then removing them if the points intersect.

\subsection{Ultrasonic Curb Detection}
For identifying curbs using the ultrasonic sensor, we use a mix of traditional filtering and statistical techniques~\cite{ultrasonicpaper}. We have 4 sensors in our test rig, and use data from all 4 of them. As in \cite{ultrasonicpaper}, we use multiple sensors to gain confidence in our measurements. This is essential when we have noisy sensors like an ultrasonic sensor. Also, using 4 sensors gives us a way to estimate the confidence of the readings at any given time. We also use the values of each sensor over past time steps and use traditional signal filtering techniques such as median filters to further reduce the noise in each signal. 
The technique in \cite{ultrasonicpaper} uses multiple values from laterally placed sensors to predict one distance measurement. This results in a delay in the output when there are changes in the curb distance, for example, during lane changes or at shoulder lanes and intersections. Using the value of the 4 different sensors separately also gives us a way to estimate bends in the curb. Hence,  in each round of measurements, we obtain 4  different values, which represent the curb distance. While we use the technique in \cite{ultrasonicpaper} to estimate the confidence associated with a set of readings, our final value for each sensor only relies on filtering that is applied over the past readings of that sensor.

\begin{figure}[]
\centerline{\includegraphics[width=1.0\columnwidth]{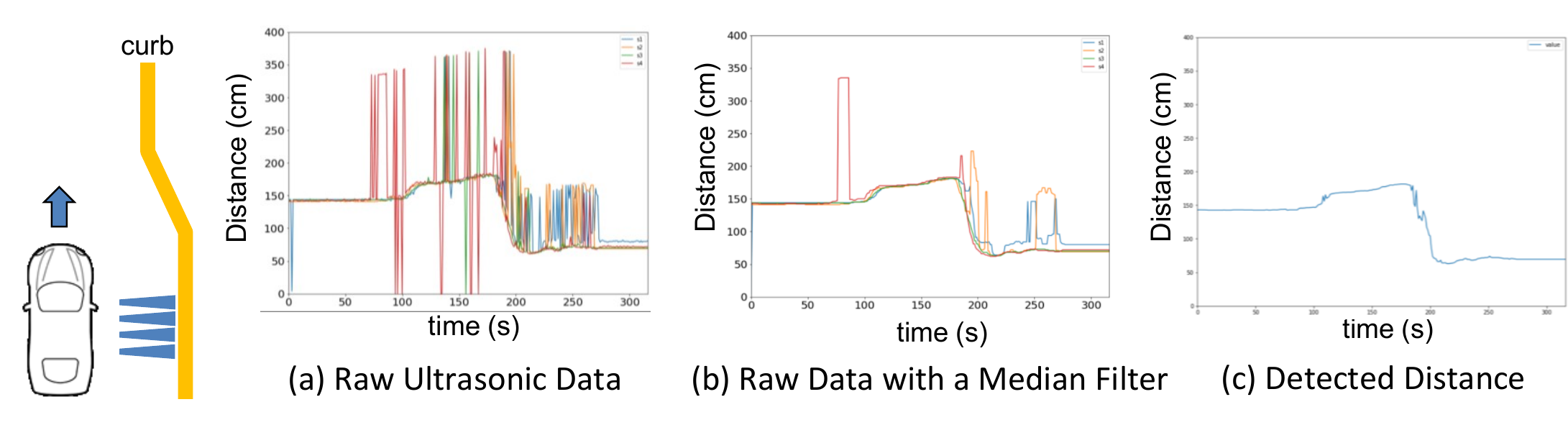}}
\caption{Left: Test driving scenario with road curb. Right: (a) Raw ultrasonic data include various noise. (b) The noise are supressed by a median filter (c) Detected distance result from the Ultrasonic-based curb detector.}
\label{fig_Ultrasonic_signal_processing}
\end{figure}

\begin{figure}[b]
\centerline{\includegraphics[width=0.85\columnwidth]{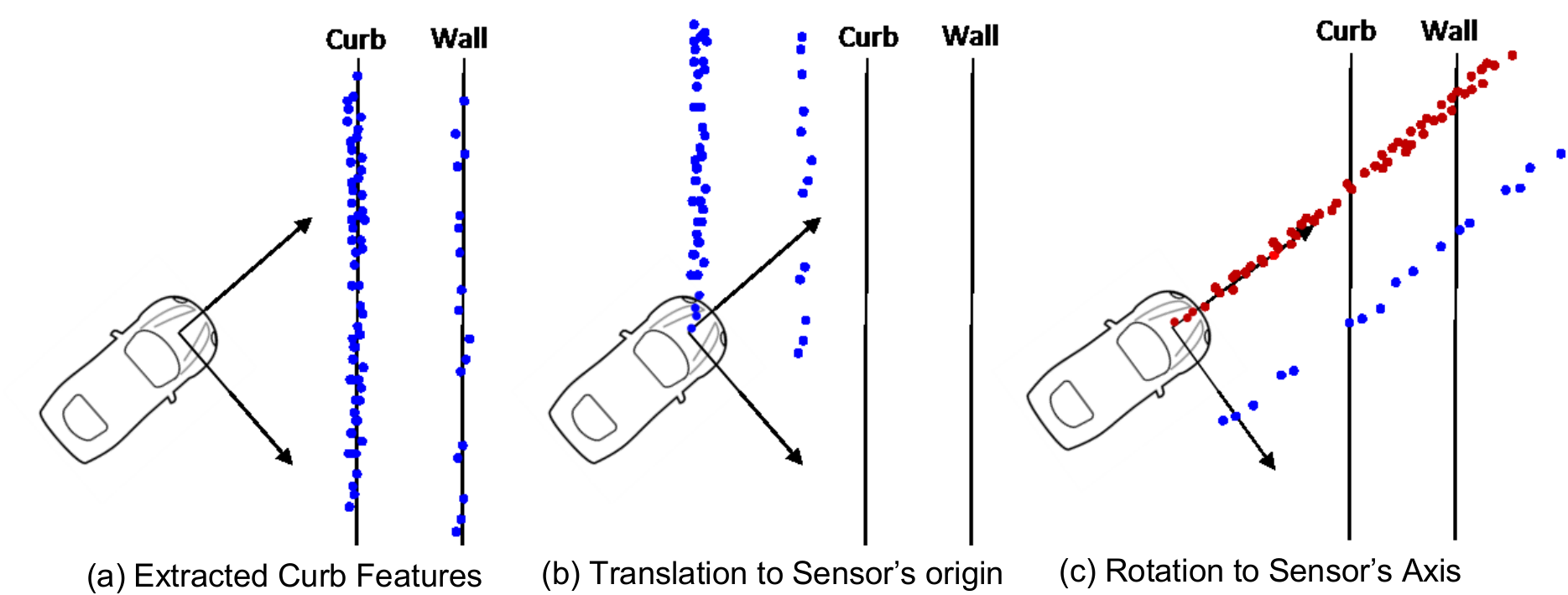}}
\caption{Proposed new curb line fitting: Theta Shift Method is deterministic, which generates the same line result using the same points and can effectively classify the curb line on complex urban roads. The blue points are extracted curb points including false positives. The red points are selected as curb points using our method.}
\label{fig_theta_shift}
\end{figure}

\begin{figure}[b]
\centerline{\includegraphics[width=0.8\columnwidth]{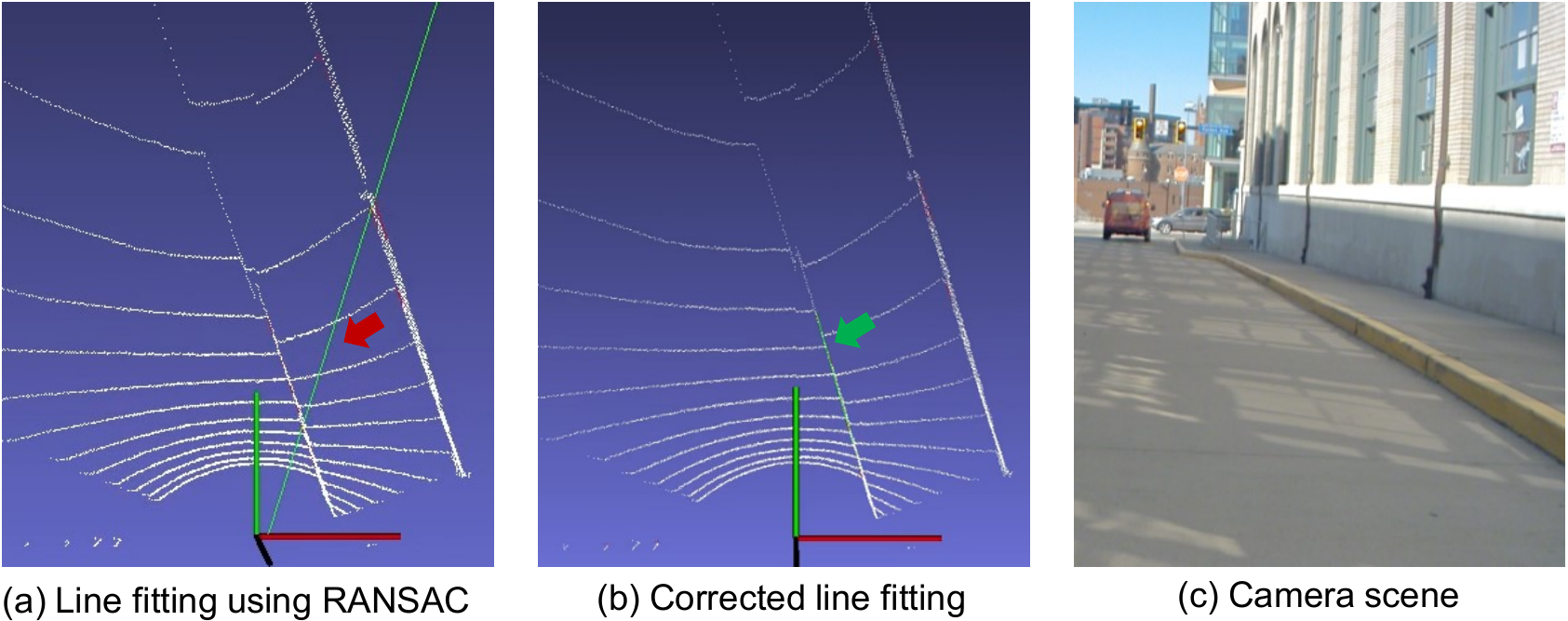}}
\caption{Example result of the Thera Shift line fitting: (a) The red points represent extracted curb points, including false positives. The green line is the calculated curb line. The red arrow indicates the wrong line fitting between curb and wall using RANSAC. (b) The green arrow shows the corrected curb line from our method.}
\label{fig_LiDAR_fitting_comparison}
\end{figure}

\smallskip
\subsection{Our Line-Fitting Approach}
\smallskip
\textbf{Theta Shift Method}

After LiDAR processing, the extracted curb features are represented by points in the point cloud that resemble the points in Figure \ref{fig_theta_shift}(a). The orientation $\theta$ of a line is only constant around the origin, but, since our data is not centered at the origin a new origin must be generated. The mean of the x-values is calculated and the closest point to the mean is chosen as the new origin. For the next frame of data, the point closest to the previous origin is found and, if the new point is greater than a certain threshold, it is chosen as the new origin. Then, all points are subtracted from the origin, shifting the data set as seen in Figure \ref{fig_theta_shift}(b). Next, the $\theta$ of all the points are calculated, and the mode is calculated. Then all points are yawed by the mode value of $\theta$ which aligins the y-axis of the car with the line seen in Figure \ref{fig_theta_shift}(c). Next, the x-value mode is calculated and the final set of chosen points is selected by points with an x-value (approximately) equal to the mode of the x-values.
The corresponding non-rotated points are then run through a least squares algorithm.



\smallskip
\textbf{Comparison to RANSAC}

RANSAC, as an outlier detection method, iteratively estimates the parameters of a mathematical model by assuming the data set is made up of a group of inliers and outliers, and then by randomly sampling some set of points and then fitting a model to that set. Several random samplings of data are taken and then the model that best fits a subset of data of the data is returned, meaning that RANSAC is not deterministic. 


The Theta Shift Method, being deterministic, also allows for ease of testing, debugging and playback as the results of the generated line will be the same every time. The most important part, as seen in Figure \ref{fig_LiDAR_fitting_comparison}(b), is that the line returned by the Theta Shift Method  more accurately fits the camera scene in Figure \ref{fig_LiDAR_fitting_comparison}(c), compared to the line returned by RANSAC seen in Figure \ref{fig_LiDAR_fitting_comparison}(a). The reason RANSAC often generates worse lines than the Theta Shift Method is due to the fact that the data set has two parallel lines. Thus, if a randomly chosen set includes two points close together on one line, and two points close together on another line, then a diagonal line can be generated between those two lines with a technically good fit level, even though it does not actually resemble the camera scene as seen in Figure \ref{fig_LiDAR_fitting_comparison}(a) and \ref{fig_LiDAR_fitting_comparison}(c).

\subsection{Curb Curve Tracking}


The curb curve generated by the line fitting over the curb detected points is prone to noise and occlusion issues between time frames.
To address this temporal incoherence problem, we smoothen the detected curve over time frames using an Extended Kalman Filter. This reduces the fluctuation of curb curves across time.

The state of the system is represented in polar coordinates. A fixed number of uniformly sampled points are selected from the detected line which is considered to be the observed states. A filter instance is then applied over each point to provide an update of the estimated state of the current time frame based on the previous one. Lastly, the coefficents of the curve are computed based on the least-squares solution line described by the updated point locations.

The curb curve prediction model extended from \cite{pclcurbdetection} is used for tracking. The two coordinate frames
$p_k = (xv_k, yv_k )$ and $p_k+1 = (xv_k+1, yv_k+1)$ represent the vehicle coordinates at
time $k$ and $k$ + 1 respectively for derivation of the curb curve-
tracking algorithm. The transformation to polar coordinates is given by 

\smallskip
\scalebox{0.75}{
\begin{minipage}{0.97\linewidth}

$$
\left(\begin{array}{c}
{\rho} \\
{\phi} \\
{\dot{\rho}}
\end{array}\right)=\left(\begin{array}{c}
{\sqrt{p^{\prime}_{{x}^{2}}+p_{y}^{\prime 2}}} \\
{\arctan \left(p_{y}^{\prime} / p_{x}^{\prime}\right)} \\
{\frac{p_{x}^{\prime} v_{x}^{\prime}+p_{y}^{\prime} v_{x}^{\prime}}{\sqrt{p_{x}^{\prime 2}+p_{y}^{\prime 2}}}}
\end{array}\right)
$$
\end{minipage}%
}

\smallskip
The state of points in polar coordinates is given by,  

\smallskip
\scalebox{0.8}{
\begin{minipage}{0.97\linewidth}

\begin{equation}x = \{\rho, \theta\}\end{equation}
\begin{equation}\rho_{t}=A \rho_{t-1} + w_{\rho,t}\end{equation}
\begin{equation}\theta_{t}=A \theta_{t-1} + B \delta \theta_{t}+ w_{\theta,t}\end{equation}
\begin{equation}\rho^{\cdot}_{t}=A \rho^{\cdot}_{t-1} +  w_{\rho^{\cdot} ,t}\end{equation}

\end{minipage}%
}

\smallskip
The prediction stage with $A$ as the state transistion matrix, $B$ as the input matrix, $P_{t}$ as the covariance ,$w_{t}$ and $Q_{t}$ as the process noise and covariance noise is given by

\smallskip
\scalebox{0.8}{
\begin{minipage}{0.97\linewidth}

\begin{equation}
\begin{array}{l}
    {\overline{x_{t}}=A x_{t-1}+B u_{t}+w_{t}}\\
    {\overline{P_{t}}=A P_{t-1} A^{T}+Q_{t}}
    \end{array}
\end{equation}

\end{minipage}%
}

\smallskip
The correction stage with $z_{t}=H x_{t}+v_{t}$ as the measurement is given by

\smallskip
\scalebox{0.8}{
\begin{minipage}{0.97\linewidth}

\begin{equation}
\begin{array}{l}
{K_{t}=\overline{P_{t}} H^{T}\left(H \overline{P_{t}} H^{T}+R_{k}\right)^{-1}} \\
{x_{t}=\overline{x_{t}}+K_{t}\left(z_{t}-H \overline{x_{t}}\right)} \\
{P_{t}=\left(I-K_{t} H\right) \overline{P_{t}}}
\end{array}
\end{equation}

\end{minipage}%
}

\smallskip
Here, $\rho$ and $\theta$ are the distance and azimuth angle of curb points respectively, and $K_t$ is the Kalman gain. The Jacobian matrix is given by

\smallskip
\scalebox{0.8}{
\begin{minipage}{0.97\linewidth}
$$
H_{j}=\left[\begin{array}{cccc}
{\frac{p_{x}}{\sqrt{p_{2}^{2}+p_{y}^{2}}}} & {\frac{p_{y}}{\sqrt{p_{z}^{2}+p_{y}^{2}}}} & {0} & {0} \\
{-\frac{p_{y}}{p_{y}\left(p_{y} p_{y}-p_{y}^{2}\right.}} & {\frac{p_{z}\left(y_{y}\right)_{z}^{2}}{\left(p_{z}^{2}+p_{y}^{2}\right)^{3}}} & {0} & {0} \\
{\frac{p_{y}\left(v_{z}^{2}+p_{y}^{2}\right)}{\left(p_{z}^{2}+p_{y}^{2}\right)^{3 / 2}}} & {\frac{p_{z}\left(y_{y} p_{z}-v_{z} p_{y}\right)}{\left(p_{z}^{2}+p_{y}^{2}\right)^{3 / 2}}} & {\frac{p_{z}}{\sqrt{p_{z}^{2}+p_{y}^{2}}}} & {\frac{p_{y}}{\sqrt{p_{z}^{2}+p_{y}^{2}}}}
\end{array}\right]
$$

\end{minipage}%
}

\smallskip
Popular tracking methods \cite{xu2016road}, \cite{zai20173}, \cite{zhang2018road} for curb lines use IMU data to calibrate Kalman filter. We do not use any IMU data instead, we use inherent distance and spatial features between points to track them in time.  Equations 4 and 5 capture the state variables. The process covariance matrix of the filter is proportional to the z-plane standard deviation of LiDAR range z distances as described in \cite{rgbd2014filter}, such that, $Q_t = \sigma^{2}_k$ where $ \sigma_k \propto d$, $d$ being the distance of the z-plane from the center, as the distance of the point. Using this relation of covariance noise and distance, the range-sampled points are tracked in time along the given path. 



\section{Evaluation}
\label{evaluation}
This section describes our experiments measuring the performance of our CurbScan system using different hardware platforms and data acquired in various road environments.

\begin{figure}[b]
\centerline{\includegraphics[width=1.0\columnwidth]{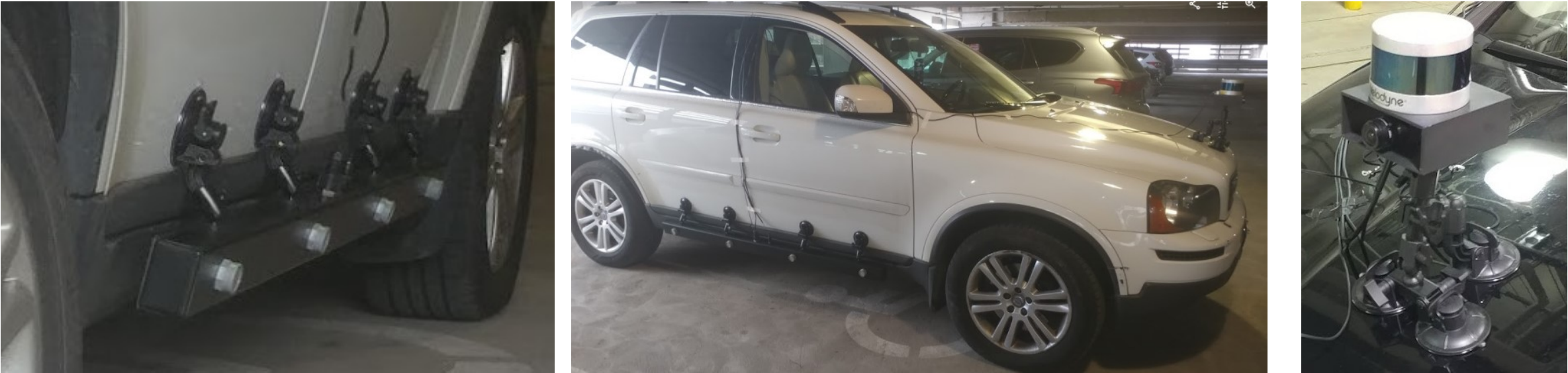}}
\caption{Left: The ultrasonic sensors are installed considering the vertical field-of-view of the sensor and average height of the curb, Middle: Test vehicle setup with sensors, Right: LiDAR and camera module are installed to minimize occlusion by the car body.}
\label{fig:test_car_setup}
\end{figure}

\begin{table}[]
	\begin{center}
		\scalebox{0.8}{%
			\begin{tabular}{|P{2.2cm}|P{3cm}|P{2.7cm}|}\hline
				Platform &\makebox[3em]{CPU}&\makebox[3em]{GPU}
				\\\hline\hline
				x86 & 3.4Ghz Intel I7 & GTX 1070 \\
				& 4 cores & 1.6Ghz 1920 Cores \\\hline
				NVIDIA Xavier & 2.2Ghz ARM v8.2  & Volta GPU \\
				& 8 cores & 1.3Ghz 512 Cores \\\hline
			\end{tabular}
		}
	\end{center}
	\caption{System Specification of different platforms}
	\label{table:platform_specs}
\end{table}

\subsection{Hardware and Implementation}
\label{hw_implementation}
In this evaluation, we examined the practical feasibility and efficacy of our algorithm on multiple platforms from different vendors. Figure~\ref{fig:test_car_setup} shows our perception system prototype to capture camera, LiDAR, and ultrasonic data and evaluate our curb detection algorithm in real-world scenarios. We use one 1.3M Pixel 60$\degree$ field-of-view (FOV) camera from FLIR~\cite{flir} and one 16-channel 360$\degree$ FOV LiDAR from Velodyne~\cite{Velodyne}.  We also integrated our algorithm on an NVIDIA Xavier~\cite{NVIDIA_Xavier} embedded platform and an x86 desktop to compare the relative performance of our detection and tracking performance across different platforms.

\begin{figure}[b]
\centerline{\includegraphics[width=0.8\columnwidth]{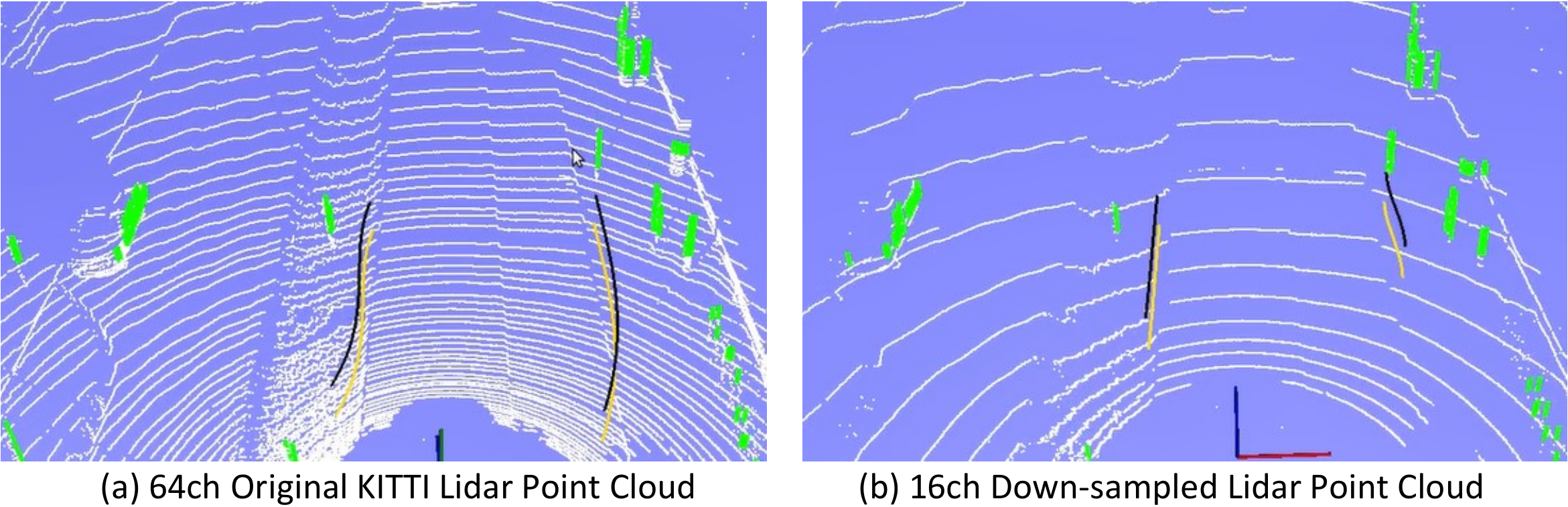}}
\caption{Test result between KITTI 64ch vs. downsampled 16ch: The yellow lines represent the detected curb lines, and the black lines are results from the tracking module. Green bars indicate detected obstacles using VScan.}
\label{fig_downsampling_test}
\end{figure}

\subsection{Experiment Results}
\label{experiment_result}

We measured the performance and accuracy for 4 different module combinations:
\begin{itemize}
    \item Experiment 1: Detection only
    \item Experiment 2: Detection and false-positive removal
    \item Experiment 3: Detection and tracking
    \item Experiment 4: Detection, tracking, and fusion with Ultrasonic sensors
\end{itemize}

In Experiments 1 and 2, we compare the improvement in accuracy from removing false positives.
Likewise, the metrics from Experiments 2 and 3 were compared to evaluate the speed and accuracy trade offs when combining the tracking module.
In Experiment 4, we demonstrate the benefit of low-cost ultrasonic sensors by fusion with lateral distance information.

\smallskip
\noindent\textbf{Dataset}

We collected LiDAR data from Velodyne VLP-16 at 10 hz and video sequences at 1280x720 resolution at 30 fps using our sensor system equipped as shown in Figure~\ref{fig:test_car_setup}.
We used 5 test sequences to create our test dataset. 
We also used the KITTI dataset~\cite{geiger2012we} for comparison purposes.
We extended an open-source tool \cite{wang2019latte} to annotate curb points. We drag over the birds-eye view of the point cloud to highlight and annotate points of interest. The annotated curb points are converted into a third-order polynomial curb model and are stored as ground truths.

\smallskip
\noindent\textbf{Downsampling}

The KITTI dataset provides point cloud data collected using Velodyne HDL-64E using 64 scan lines. The Velodyne scans are stored as floating-point values and {x, y, z, reflectance} for each point. However, it does not come with the ring number, or scan number, that each point belongs to from the data set. Since the scan number information is necessary information to run our algorithm, we implemented a tool to add this information to the data. We extended our method to downsample the point cloud from 64 scans to 32/16 scans. 
The points from the same scan have the same fixed theta value, which is provided in the Velodyne manual~\cite{Velodyne16Manual}. However, fr theta values from the same scan inevitably have some errors. To find the points within the same scan, we first run a clustering algorithm on the test data to get an average theta value and the estimated range for each scan. These values are then used to classify the points to 64 scans. Once the points are classified, we downsample the point cloud by choosing the scan number we need. We downsampled them to 16 scans for fair comparison with VLP-16.

\begin{figure}[]
\centerline{\includegraphics[width=1.0\columnwidth]{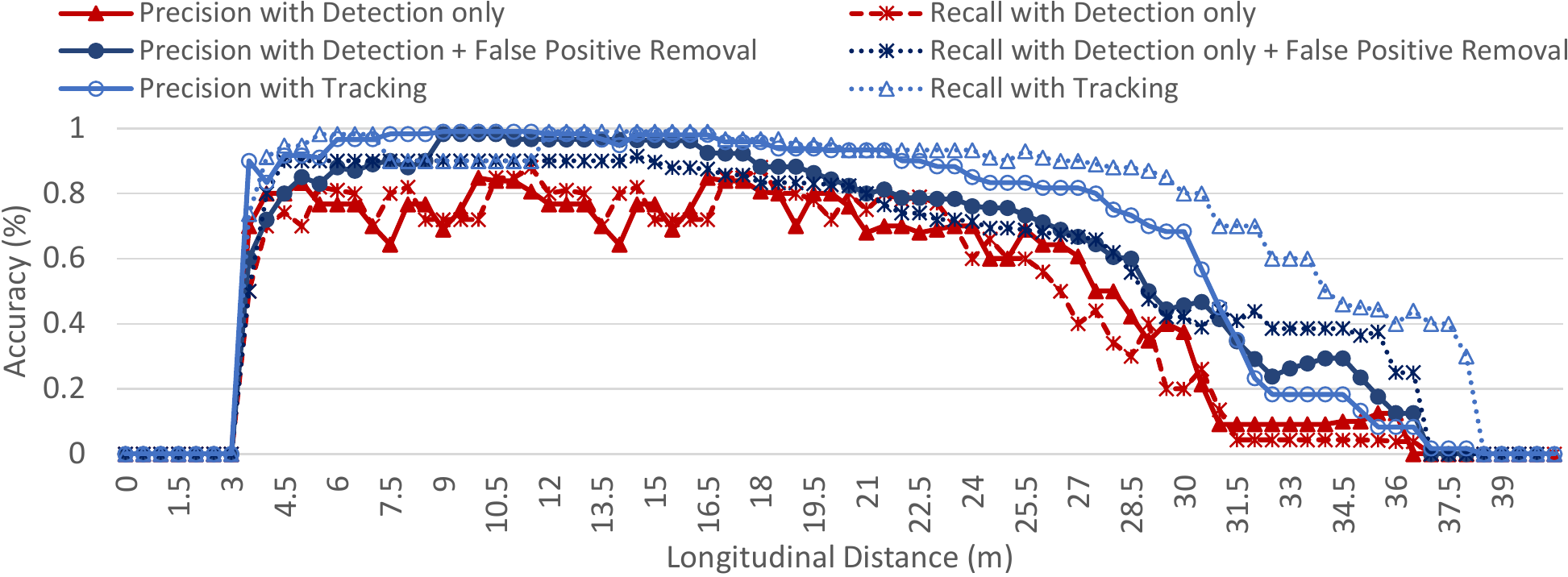}}
\caption{Accuracy comparison using KITTI downsampled to 16 scans}
\label{fig_accuracy_comparison_kitti}
\end{figure}

\begin{figure}[]
\centerline{\includegraphics[width=1.0\columnwidth]{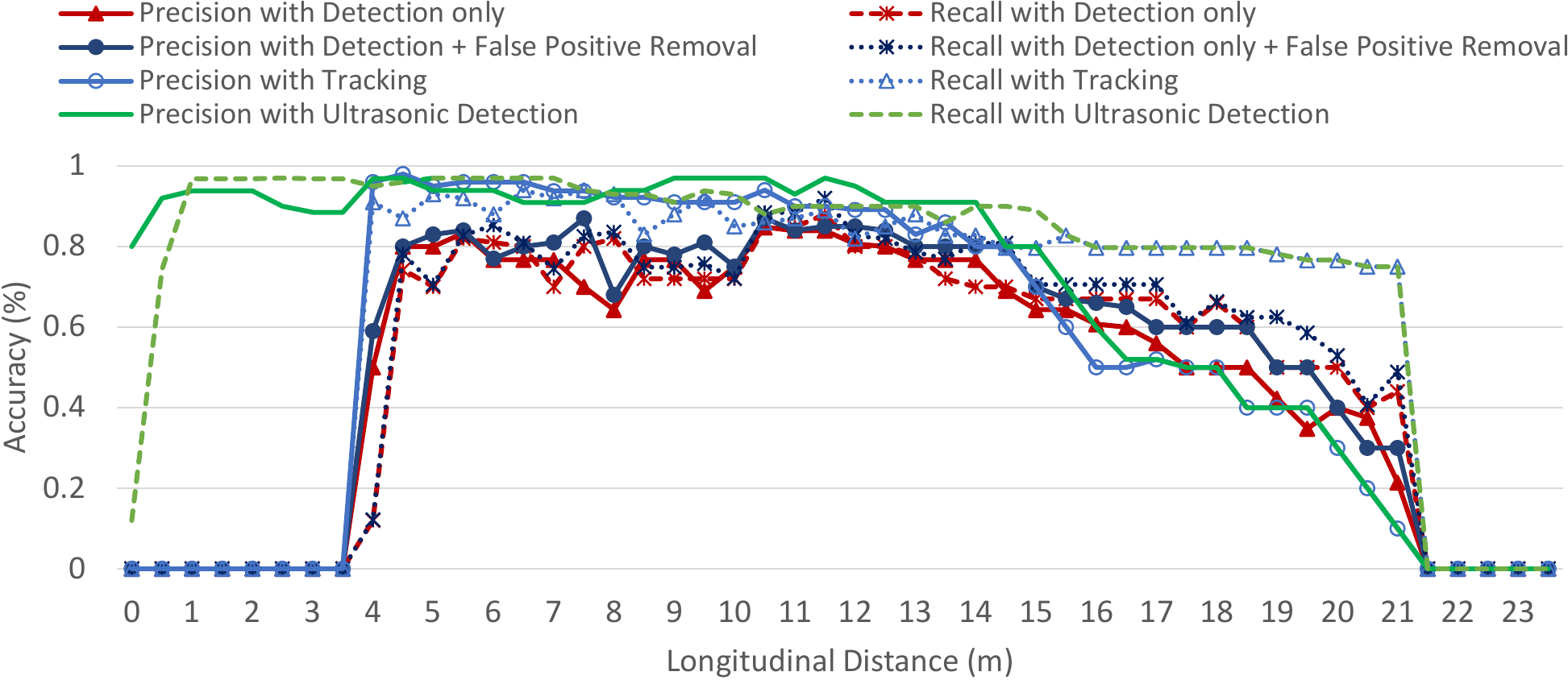}}
\caption{Accuracy comparison using Our Dataset with VLP-16}
\label{fig_accuracy_comparison_vlp16}
\end{figure}

\smallskip
\noindent\textbf{Third-Order Polynomial Curb Model}

We chose third-order polynomials, which sufficiently describe the curbs within the effective range of VLP-16, as the output from our algorithm. In our LiDAR configuration with the y-axis towards the vehicle front and x-axis towards the right of the vehicle, we define Y as the longitudinal distance and X as the lateral distance from our LiDAR sensor. The lateral distance from the longitudinal axis is 

\scalebox{0.8}{
\begin{minipage}{0.97\linewidth}
\begin{equation}
X = C3 \cdot Y^3 + C2 \cdot Y^2 + C1 \cdot Y + C0
\end{equation} 
\end{minipage}%
}

\smallskip
The outputs are coefficients from $C3$ to $C0$ and [$Y_{min}$, $Y_{max}$], which describe the range of detection.

\smallskip
\noindent\textbf{Accuracy Calculation}

To evaluate the accuracy of our algorithms, we used the following metrics for detection and tracking modules.

\scalebox{0.8}{%
$precision=\frac{Correctly\ detected\ Curb\ lines (True\ Positive)}{All\ Detected\ Curb\ lines
(True\ Positive + False\ Positive)}$
}
\newline

\scalebox{0.9}{%
$recall= \frac{Correctly\ detected\ Curb\ lines (True Positive)}{(True\ Positive + False\ Negative)}$
}

\smallskip
Besides, we evaluate the accuracy considering the longitudinal distance in front of the car. The accuracy per the longitudinal distance is critical because it can provide precise confidence per the distance to the localization module in autonomous vehicles.
We compare test results and the ground truth, both represented as polynomials, in every interval within the range. By sampling points in Y, the corresponding X, which are the lateral distances, are compared and classified into four categories: \emph{True Positive}, \emph{False Positive}, \emph{True Negative}, \emph{False Negative}. Then, precision and recall are calculated in every defined interval as shown in Figure~\ref{fig_accuracy_comparison_kitti} and \ref{fig_accuracy_comparison_vlp16}. 
As in the figures, the addition of the false-positive removal module improves both precision and recall by reducing false positives and increasing true positives.
The tracking module improves the recall metric because it reduces false-negatives caused by occluded curb points due to nearby obstacles.
However, the tracked precision results after 31m for the KITTI and 15m for our dataset decrease because the sparse points cause unstable prediction.
Our algorithm with the tracking module shows good accuracy within 20 meters and 14 meters for the KITTI dataset and our dataset, respectively. The KITTI data shows better longitudinal accuracy because their LiDAR is mounted on the top of the vehicle, which allows a farther viewpoint.
We observe that no detection was found in the close area, 0 and 4 meters, to the car using the LiDAR-only solution because the ego-vehicle occludes the range. The ultrasonic sensor can be beneficial in this case. The accuracy between 0 and 4 meters can significantly be improved since the ultrasonic sensors provide lateral distance from the side of the car to the curb. The ultrasonic sensors can also increase overall accuracy over the entire longitudinal ranges in case the curb is close to the vehicle.  However,  we observed noisy information after 7 meters from the ultrasonic sensors. Our algorithm, therefore, trusts the distance information from the ultrasonic sensors only when they provide constant distances during a specified interval.

\begin{table}[]
	\begin{center}
		\scalebox{0.9}{%
			\begin{tabular}{|P{2.5cm}|P{1.7cm}|P{1.7cm}|P{1.7cm}|}\hline
				\backslashbox[30mm]{Platform}{Module}
				&\makebox[5em]{LiDAR Detector}&\makebox[5em]{With Tracking}&\makebox[5em]{With Ultrasonic}
				\\\hline\hline
				x86 & 100 fps & 84 fps & 100 fps \\\hline
				NVIDIA Xavier & 10 fps & 8 fps & 9 fps \\\hline
			\end{tabular}
		}
	\end{center}
	\caption{End-to-end System Performance}
	\label{table:table_speed}
\end{table}

\subsection{CurbScan Latency Analysis}
\label{experiment_result}
In this section, we evaluate the speed of the algorithm on an x86 machine and an NVIDIA Jetson AGX Xavier. 
Table II summarizes the latency in fps on the x86 and Xavier with and without tracking and the ultrasonic sensors.


The LiDAR detection algorithm runs sequentially and is much faster on the x86. With a many-core, GPU integrated computer like the Xavier, the algorithm can benefit significantly from parallel optimizations. The latencies may drop further for both architectures, once the algorithm is optimized. The latency for the ultrasonic sensors which runs in parallel with the Lidar detection algorithm was also measured. 
The ultrasonic sensor processing is much faster on the Xavier, due to the libraries used to compile the code, with the Xavier having the advantage of optimized \emph{scipy} and \emph{pandas} libraries. Since the ultrasonic sensors run at 40Hz compared to the Lidar detection's speed of 10Hz, the LiDAR detection is not slowed down by the ultrasonic sensors. 

\section{Conclusions and Future Work}

In this paper, we presented a real-time solution that can detect and track curb lines using multiple sensors: sparse LiDAR data, a mono camera, and low-cost ultrasonic sensors.
In particular, we adopted the multiple filter method to extract candidate curb points and developed a practical masking method to select correct curb points.
We additionally proposed a new regression algorithm to provide robust curb line fitting results. We also demonstrated the benefit of low-cost Ultrasonic sensors by fusion with lateral distance information.
Our CurbScan system was evaluated on x86 and NVIDIA Xavier platforms, and their relative performance was measured for various combinations of active modules. The final solution was also deployed on a real car and tested in real-world situations. The system performed with good accuracy and precision in these real-world tests and other open datasets such as KITTI.
As a next step, it would be interesting to expand our solution to a system with multiple LiDARs to deal with the intersections and broad road areas.
Another opportunity is to expand our algorithm to detect general road boundaries: wall, fence, speed bump, parked cars, etc.

\bibliographystyle{IEEEtran} 
\bibliography{main} 

	


\end{document}